\documentclass[journal]{IEEEtran}

\usepackage{hyperref}

\usepackage{times}
\usepackage{epsfig}

\usepackage{graphicx}

\usepackage{amsmath}
\usepackage{amssymb}

\usepackage[normalem]{ulem}

\DeclareMathOperator*{\argmin}{arg\,min}
\DeclareMathOperator*{\argmax}{arg\,max}

\usepackage{array}
\usepackage{algorithm, algpseudocode}

\usepackage{xcolor}

\usepackage[utf8]{inputenc}

\usepackage[switch]{lineno}
\newcommand{\revised}[1]{\textcolor{black}{#1}}
\definecolor{replygreen}{rgb}{0, 0, 0}

\renewcommand{\sout}[1]{\unskip}

\ifCLASSINFOpdf
\else
\fi

\begin{document}

\title{Real-time high speed motion prediction using fast aperture-robust event-driven visual flow}

%
%
\author{Himanshu Akolkar,
        Sio-Hoi Ieng,
        and~Ryad~Benosman
\thanks{Himanshu Akolkar is at University of Pittsburgh, USA}
\thanks{Ryad Benosman is at Robotics Institute, Carnegie Mellon University, USA, University of Pittsburgh, USA and Sorbonne Universite, INSERM, CNRS, Institut de la Vision, France}
\thanks{Sio-Hoi Ieng is at Sorbonne Universite, INSERM, CNRS, Institut de la Vision, France}}

\maketitle

\begin{abstract}
Optical flow is a crucial component of the feature space for early visual processing of dynamic scenes especially in new applications such as self-driving vehicles, drones and autonomous robots. 
The dynamic vision sensors are well suited for such applications because of their asynchronous, sparse and temporally precise representation of the visual dynamics. Many algorithms proposed for computing visual flow  for these sensors suffer from the aperture problem as the direction of the estimated flow is governed by the curvature of the object rather than the true motion direction. Some methods that do overcome this problem by temporal windowing under-utilize the true precise temporal nature of the dynamic sensors. In this paper, we propose a novel multi-scale plane fitting based visual flow algorithm that is robust to the aperture problem and also computationally fast and efficient. Our algorithm performs well in many scenarios ranging from fixed camera recording simple geometric shapes to real world scenarios such as camera mounted on a moving car and can successfully perform event-by-event motion estimation of objects in the scene to allow for predictions of upto 500~ms i.e. equivalent to 10 to 25 frames with traditional cameras. \end{abstract}


%

\section{Introduction}
%
%
%
%
\IEEEPARstart{O}{}ptical flow is the measure of motion of an object projected on to the image plane of a camera. It is one of the fundamental steps needed for understanding a dynamic visual scene and has taken an even important role with newer applications such as autonomous driving vehicles \cite{Menze_2015}, drones, action perception during user interactions \cite{Simonyan_2014} in robots and traditional applications like video editing \cite{Bonneel_2015} and stabilization. Because the visual sensing has traditionally been based on image acquisition at fixed time intervals, the computation of optical flow has been based on finding features that move across two or more consecutive images. Since the intensity of light received on the sensor is the most basic feature, the first principle approach for measurement of optical flow is given by the `brightness constancy assumption' that assumes that the brightness of an object moving across the camera remains constant over short internal of time. Ideally this time interval should be infinitesimal, but practically, for the traditional cameras, this means the time between two recorded frames. This constant instantaneous brightness assumption forms the basis for the earliest algorithms such as those proposed by Horn and Schunk \cite{HornSchunk} and the Lucas-Kanade (LK) algorithm \cite{LucasKanade}. This has been further expanded to `constant feature assumption' where complex features or descriptors are extracted \cite{Wang_94} and tracked over multiple spatial scales \cite{Brox_2009}. With the advancements in convolution and deep neural networks, a number of new algorithms using these approaches have been proposed to compute visual flow \cite{ Brox_2004, Bai_2016, Bailer_2017}. Some of these methods even propose tackling optical flow computation as a learning problem \cite{Brox_2016}. While these approaches intend to achieve high accuracy using the improving computational power of GPUs and FPGAs, the fundamental problem of fast sensing and image processing still poses a hinderance towards using such techniques as part of a larger perceptive autonomous system. \\
The new generation of dynamic visual sensors \cite{DVS, Bernabe, ATIS} might be able to fill in this niche application space by virtue of their fast, accurate sensing of light with high temporal precision. In this paper, we propose an algorithm designed for use with one such type of sensor \cite{ATIS}. Event-driven sensors have evolved over the last few years as possible successors to frame based classical cameras, especially for visual sensing in research areas that require high precision over a large temporal dynamics range like robotics \cite{Grover_2016, Akolkar_2015, Vasco_2016, Rea_2013}, autonomous vehicles \cite{Scaramuzza_2018_1} and navigation in drones \cite{Scaramuzza_2018_2}. As these sensors provide precise motion information due to the inherent design of the pixels, they are ideal for fast visual flow computations. \\
A number of methods have been proposed to compute visual flow using event based sensors. As events in the event-driven sensors are essentially encoding the light intensity captured by the pixels, algorithms based on the original image based Lucas-Kanade method have been proposed \cite{Benosman2012}. While these event-driven derivatives are fast, they cannot achieve the same accuracies as the frame-based variants due to the loss of information in conversion from intensity to events.\\
Several algorithms are designed specifically to take advantage of the temporal nature of event-driven paradigm \cite{Benosman2014,Brosch2015}. These algorithms use the spatio-temporal structure of events to estimate the flow by fitting a surface (usually a plane) and compute the normal of this surface as flow estimate. These algorithms maybe classified under the label of `plane-fitting algorithms'. While these algorithms have improved accuracy of event flow, they are limited to computations of local dense flow. Further, the flow obtained is always computed as orthogonal to the edge irrespective of the direction of true motion. Thus, the flow computation is susceptible to the gradient of the edge. This problem is referred to as the aperture problem. The only way to tackle the aperture problem with a traditional plane-fitting method is to increase the size of the spatial neighborhood around the events when fitting the plane but this can lead to errors as the true size of the object is unknown and the shape of object {might not remain} linear. \\
A recent algorithm has been able to avoid this problem using constrained statistical properties of the object but it is computationally too intense to be used in real time and is only valid for object with closed form \cite{Seifozzakerini2017}. Another recent method for computing event-driven visual flow uses a spatio-temporal window of events and performs histogram matching of the event clusters to estimate the direction and speed of object. Thus, the current state-of-the-art algorithms lose the temporal dynamics of the input sensor events as they require pooling of events over a temporal and spatial window to avoid aperture problem. \\
Here we propose a new event-driven algorithm to solve aperture problem using multi-scale spatial pooling that uses the local erroneous flows computed at the lowest scales and corrects their direction towards the true direction of motion of the object. We mathematically prove that because of the specific properties of the plane fitting algorithms, pooling the fast but erroneous local flows over an appropriate spatial scale can correctly estimate the real direction of the object. Further, the estimation of this spatial scale can be computed at every event independently without any a-priori knowledge about the shape and size of the objects in the scene and is independent of any global motion of the camera. 
The proposed algorithm can perform in myriad of scenarios. Finally, this flow rectification allows us to perform very low-level event predictions i.e. when and where should new events appear according to the observations. We show via experiments that we can estimate on the fly, locations and velocities of moving objects \revised{of up to} 500ms ahead in the future. Such prediction can be implemented for solving visual tasks such as collision avoidance and tracking.\\



 

\section{Methods} \label{section: methods}

\begin{figure*}[ht] 
\begin{center}
\includegraphics[width=0.7\textwidth]{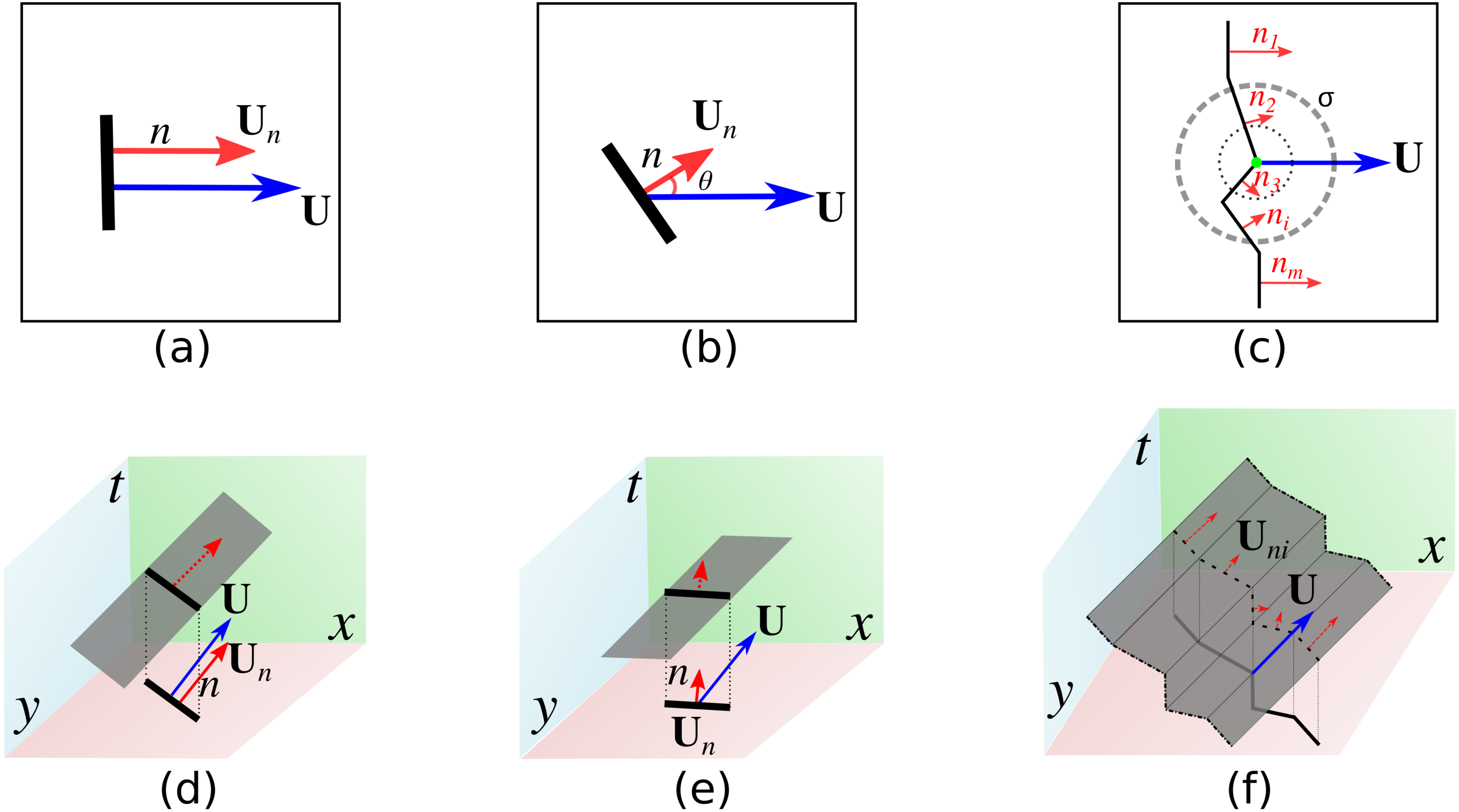}

\caption{\textbf{(a)} and \textbf{(b)} show an oriented edge moving across the sensor in true direction $\mathbf{U}$ and the predicted local flow $\mathbf{U}^T\mathbf{n}$ by fitting the plane over events in [x, y, t] space as in \textbf{(d)} and \textbf{(e)}. The magnitude of the normal velocity component estimated by the plane fitting method is related to the orientation of the edge and true motion direction as $\mathbf{U}^T\mathbf{n}= |\mathbf{U}| \cos{(\theta)}$.  This relationship can be extended to a larger complex shaped object by linearizing it using multiple small edges \textbf{(c)} over small spatial region and performing plane fitting over each local edge \textbf{(f)}. \textbf{(c)} The \revised{best estimate of} true flow direction can be estimated by finding the correct spatial size \textbf{$\sigma$} corresponding to the maximum mean magnitude $|\overline{\mathbf{U}_n}|$.} \label{fig:math_explain}  
\end{center}
\end{figure*}


The algorithm proposed in this paper uses multi-scale pooling found in biological visual system in higher animals for hierarchical object recognition. The basic idea here is to perform local, fast flow measurements which might be incorrect in their direction estimations but are relatively reliable in amplitude estimates and then correct the direction estimates using global amplitude information. 

\subsection{Multiscale pooling}



Figure~\ref{fig:math_explain} shows the principle idea motivating the correction procedure explained in the next section. Let us assume the most ideal case suited for the plane fitting method: a single bar moving in front of the camera generating a perfect event plane in the [x, y, t] space. If the bar is oriented orthogonal to its direction of motion (   Figure~\ref{fig:math_explain}(a)), the estimate of the velocity computed using the plane fitting method \cite{Benosman2014} (Figure~\ref{fig:math_explain}(d)) would be equal to the true velocity $\mathbf{U}$. But, if the bar is now rotated (b) by an angle $\theta$, the velocity estimate of the flow from plane fitting is $\mathbf{U}_n = \mathbf{U}^T \mathbf{n}.\mathbf{n}$, $\mathbf{n}$ being the unit normal to the bar. 
The signed magnitude of this flow can be given by: 

\begin{equation}
\mathbf{U}^T \mathbf{n} = |\mathbf{U}| \cos{(\theta)},
\label{eq:projection}
\end{equation}

This shows that the plane-fitting based estimated flow is equal to the true direction of motion when the magnitude of the estimate is maximum, i.e. the cosine is maximum in~Eq.~(\ref{eq:projection}).\\
{It is important to note that the normal $\mathbf{n}$, without additional assumption, can have two directions - namely either one of the two directions along the line orthogonal to the bar. However, if we are considering the temporal surface defined in~\cite{Benosman2014} (as shown in figure \ref{fig:math_explain}(d-f)) as a bi-dimensional function $t$ of $(x,y)$, where the gradient of $t$ allows us to define $\mathbf{n}$ as its unit direction vector then $t$ is always increasing in the direction of the motion (i.e. the directional derivative of $t$ along $\mathbf{U}$ is increasing) and we always have $\theta \in [-\frac{\pi}{2},~\frac{\pi}{2}]$ or equivalently $\mathbf{U}^t.\mathbf{n} \geq 0$.}



We can generalize the observation in Eq (\ref{eq:projection}) to more complex objects using this property of the plane-fitting flow computation. Figure~\ref{fig:math_explain}(c and f) shows one such example case: let us consider a contour of a random shape moving with velocity $\mathbf{U}$. We can approximate this shape as a set of line segments. For each pixel/event of each segment, the plane fitting method is estimating $\mathbf{U}_n$. If we consider a spatial neighborhood $\sigma$ around a random pixel (example : green dot in (c) ) - for which we have estimated its normal velocity, the mean speed (i.e. the amplitude of the mean velocity) computed within $\sigma$ is defined as:
\begin{equation}
|\overline{\mathbf{U}_n}|=\frac{\sum_{i\in \sigma} K_i \mathbf{U}^T\mathbf{n_i}}{ \sum_{i\in \sigma}K_i},
\label{eq:meanspeed}
\end{equation}
where $K_i$ is the length of the $i^{th}$ segment in pixels within $\sigma$ and with the assumption that all the pixels are contributing in the mean flow estimation.

If we assume that within this spatial neighborhood there lies a line segment $j$ such that it is oriented relatively closest to the true motion direction (i.e. $\theta$ is minimal and ideally $\theta_j=0$ when it is oriented orthogonal to the true velocity), we can find an upper bound for $|\overline{\mathbf{U}_n}|$: 

\begin{equation}
|\overline{\mathbf{U}_n}| \le  
\frac{\sum_{i\in \sigma} K_i \mathbf{U}^T\mathbf{n_j}}{ \sum_{i\in \sigma}K_i}
= \mathbf{U}^T\mathbf{n_j}=U_m.
\label{eq:upperbound}
\end{equation}

Since the local mean speed is upper bounded by the amplitude of the velocity that is "most" colinear to $\mathbf{U}$, the larger the $|\overline{\mathbf{U}_n}|$ we get from a given $\sigma$, the closer we are to $\mathbf{U}$ i.e. the $\sigma$ leading to the largest $|\overline{\mathbf{U}_n}|$ is the "right spatial scale" for which $|\mathbf{U}-\overline{\mathbf{U}_n}|$ is minimized. As we do not know the true velocity $\mathbf{U}$, $U_m$ becomes our next best reference for \textbf{U}.\\
According to this observation, for a given flow estimate, we define the problem of correction as the minimization problem of finding the neighborhood scale, $\sigma$, for which the cluster of flow estimates whose mean magnitude is close to the theoretical maximum $|\mathbf{U}| \approx U_m$ as described previously. Thus, for given neighborhood $\sigma$, we define the error function as:

\begin{equation}
E_{\sigma} = \frac{\sum_{i\in \sigma}K_i(U_m - \mathbf{U}^T\mathbf{n}_i)}{\sum_{i\in\sigma} K_i}.
\label{eq:minimize}      
\end{equation}
We then have according to~(\ref{eq:upperbound}):
\begin{equation}
	E_{\sigma} = U_m - |\overline{\mathbf{U}_n}|.
	\label{eq:error}
\end{equation}

Since $0 \le |\mathbf{U}_n| \le U_m$, the problem of finding the right $\sigma$ is equivalent to the minimization problem 
\begin{equation}
\begin{split}
\argmin_{\sigma}(E) = \argmin_{\sigma} (U_m - |\overline{\mathbf{U}_n}|) 
\\\equiv \argmax_{\sigma} (|\overline{\mathbf{U}_n}|).
\\\equiv \argmin_{\sigma} (|\overline{\mathbf{\theta}}|).
\end{split}
\label{eq:optimize}
\end{equation}

The above equations show that finding the scale with maximum mean magnitude is equivalent to finding the scale which best estimates the direction of true global flow. \revised{Eq. \ref{eq:error} and \ref{eq:optimize} combine to give the new estimated flow, magnitude and direction, from the optimal spatial scale.}
Thus, we only require to compute the flow over the smallest scale once, and perform the above maximization over larger spatial scales to get the \revised{best estimate of the} true global motion direction. \revised{Since, the above method is an optimization problem, the resulting estimate of the direction depends on the available line segments directions in the scene. While the method can give us the "true" direction of the object, in non-ideal conditions when the line segment that is orthogonal to the true direction is missing, the method can only provide the closest best estimate of the true direction. Further, as the estimated flow is given by the average local flows in the optimal spatial scale, it is possible that the flow is slightly biased by the flow values from the incorrect orientations. Since the incorrect flow values decrease with cosine of the angle, this bias is generally very small - nonetheless, this means that rather than getting the exact true flow, we will get very small errors in the flow.} The proposed algorithm can therefore be divided into three steps. First, we compute local flow for each event using plane fitting. Second, we search for a spatial scale for which the mean magnitude of these local flows is maximized. Third, we calculate the mean direction for the flows in this scale and assign the direction to  all the local flow events within this scale. 

{Interestingly, while we define Eq.~(\ref{eq:projection}), in regards to the plane fitting algorithm, this property holds true for many other methods of computing local flow such as the Lucas-Kanade flow \cite{LucasKanade} or their event based versions \cite{Benosman2012}. This means that while this paper in the following section uses plane-fitting to compute and correct local flow, the proof below shows that any existing flow methods that adhere to the relationship in Eq.~(\ref{eq:projection}) can be corrected using the multi-scale correction method.}

\section{{Algorithm Implementation}} \label{section: algorithm}

\begin{algorithm}[ht]
\begin{algorithmic}[1]
\caption{Multi-scale aperture robust optical flow}
\For{each event {x, y, t} } \label{Event for loop}
	\State{\textbf{1. COMPUTE LOCAL FLOW (EDL):}} 
    \State{Apply the plane fitting ~\cite{Benosman2014} to estimate the plane parameters [a, b, c] within a neighborhood {(5x5)} of $(x,y,t)$.}
    \State{Set $\hat{U} = ||(a, b)||$
    		and Inliers\_count = 0}
    \State{$\hat{{z}} = \sqrt{a^2 + b^2}$}
    		\For{each {event} $(x_i, y_i, {t_i})$ in neighborhood N{=5}}
    		\State{$\hat{t} = (ax_i-x) + (by_i - y)$}
            	\If{$|t_i-\hat{t}| < \frac{\hat{{z}}}{2}$}
            		\State{Inliers\_count = Inliers\_count+1}
            		\EndIf
            \EndFor \label{inlier_check}
     		\If{Inliers\_count $\geq$ $0.5*N^2$}
            \State{Set $\theta=arctan(a/b)$ and 
            $\mathbf{U}_n=(\hat{U}, \theta)^T$}
            \Else 
            	\State{$\mathbf{U}_n = (0,0)^T$.}
            \EndIf
\State{\textbf{2. MULTI-SPATIAL SCALE MAX-POOLING:}}
\State{Define $S=\{\sigma_k\}$, the set of neighborhoods, centered on $(x,y,t)$, $\sigma_k$ with increasing radius and $\delta$ t($\sigma_k$) $\leq t_{past}$ {($t_{past}$ is temporal cut-off delta)} }
	\If{$\mathbf{U}_n \neq (0, 0)^T$}
    	\For{each $\sigma_k \in S$ }
        	\State{$\overline{\mathbf{U}}_{n,\sigma_k} = \underset{j \in \sigma_k}{\text{mean}}(\mathbf{U}_{n_j})=(\overline{\hat{U}}_k,\overline{\theta}_k)^T$}
            
        \EndFor
        \State{$\sigma_{max} = \underset{\sigma_k \in S}{\text{argmax}}(\overline{\hat{U}}_k$)}
        \EndIf 
\State{\textbf{3. UPDATE FLOW:}}        
        \State{Flow~(x,y) = $=
        \underset{j\in \sigma_{max}}{\text{mean}} 
        \begin{pmatrix} 
        \hat{U}_j\cos{\theta_j}\\
        \hat{U}_j\sin{\theta_j}
        \end{pmatrix}$}
\EndFor 
\label{algorithm:main}
\State{Refer Table \ref{table: params} for parameter values.}
\end{algorithmic}
\end{algorithm}

The steps involved in the implementation of the flow are described in Algorithm~\ref{algorithm:main}. The local flow was computed using an iterative implementation of the plane fitting flow as in~\cite{David2018}. Some minor changes are introduced to the original implementation in ~\cite{David2018} to improve performance. Firstly, to improve the accuracy of the flow and remove noise, we add an error correction step to ensure better accuracy of the plane fitting by computing the number of inliers (events that are within a certain distance from the fitted plane). If the number of inliers is more than half the total points used to fit the plane, we consider the fitting to be good and the flow estimate to be reliable. This improves overall efficiency and noise robustness as the rectification is only performed on valid flow events.\\
To further avoid older events from corrupting flow estimates, we added a temporal history limit such that the correction was performed using events that occurred within a certain time ($t_{past}$) from current event. Table \ref{table: params} lists the parameters values used to estimate flow for datasets used the experiments and results in Section \ref{section: experiments}.


\begin{table}[ht]
\centering
  \begin{tabular*}{\columnwidth}{@{\extracolsep{\fill}}|c|c|}
  \hline 
      \textbf{Parameter} & \textbf{Value} \\
      \hline 
      \noalign{\vskip 2mm} 
     
      \multicolumn{2}{c}{\textit{Local flow}}\\ \hline
      Filter size $N$ & $5$ pixels \\
      \hline 
      Inlier percentage & $50\%$ \\
      \hline         
      \noalign{\vskip 2mm} 
      
      \multicolumn{2}{c}{\textit{Rectification}}\\
      \hline 
      Spatial range $\sigma$ & $0$ to $100$ pixels in steps of 10 \\
      \hline
      Temporal limit $t_{past}$ & $5$ msec \\
      \hline
  \end{tabular*}
  \caption{Algorithm parameters \label{table: params}}
\end{table}


\section{Experiments and results} \label{section: experiments}

The performance of the algorithm was measured in different scenarios to test its effectiveness over the plane fitting method. The results are divided into two sections - first, we show in four different scenarios how our algorithm corrects the direction errors over the plane fitting algorithm. In the later section we show how this corrected flow can be used to implement event-by-event predictions of moving objects. For the sake of brevity, in the rest of the text, we abbreviate the local plane fitting flow as EDL (Event Driven Local) flow while the corrected flow estimates using our algorithm as ARMS (Aperture Robust MultiScale) flow.

\subsection{Flow correction}
\subsubsection{Camera fixed, trivial pattern} \label{Exp1}
\begin{figure} 
  \begin{center}
  
   \includegraphics[width=\columnwidth]{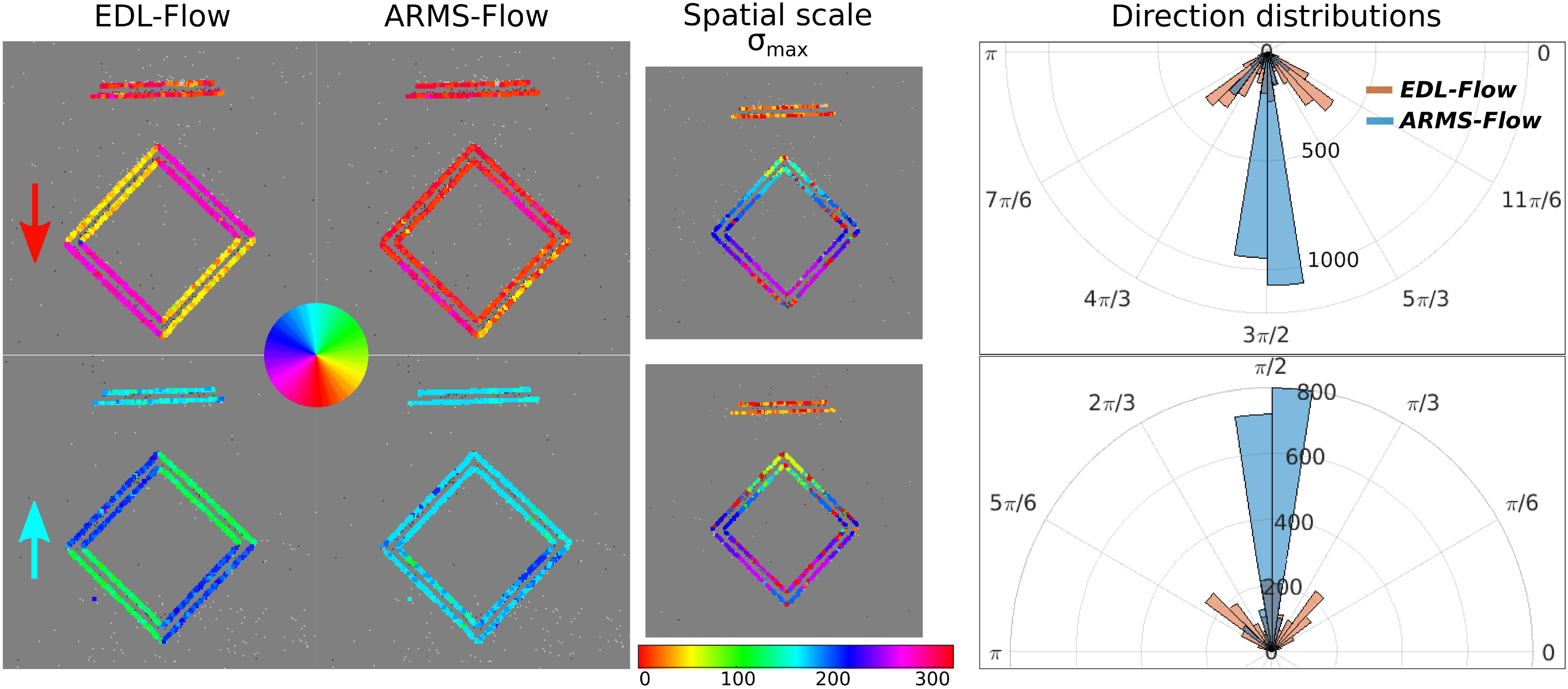}  
     
\caption{The figure shows the output of the algorithm for trivial case of  bars and squares moving up and down. (A) The direction of EDL flow estimates is normal to the edge orientations which is corrected by ARMS. \revised{(B) shows events color coded by the size of optimal window.  (C) represents the direction distributions showing} how the EDL gives three distinct peaks for each of the orientations which ARMS corrects \revised{towards a single peak representing vertical motion.} \label{fig:Fig_results_exp_1}}  
  \end{center}
  
\end{figure}

We used a simple geometric pattern of bars and squares moving up and down in front of the sensor. Figure~\ref{fig:Fig_results_exp_1}(A)[EDL Flow] shows the flow computed using just plane fitting algorithm on a given slice of events. As evident from the figure, while most of the events have correct flow direction on the bars, the flow directions of the edges of the square are incorrectly pointing towards the normal of the edges. Figure \ref{fig:Fig_results_exp_1}(A)[ARMS Flow] shows the output of our algorithm. The directions of the edges are corrected uniformly towards the true direction of motion. The quantification of these results are shown in the histograms of Figure \ref{fig:Fig_results_exp_1}[right]. The graphs in red show the distribution of directions (in radian) estimated by the plane fitting algorithm. The graph indicates tri-modal distribution for downward/upward ($\pi/2$, $3\pi/2$) and the directions along the normal to the edges ($\pi/4$, $3\pi/4$ for up and $5\pi/4$, $7\pi/4$) while the distribution of the corrected flow directions (blue) largely make up a single peak in the direction of real motion. \revised{Figure \ref{fig:Fig_results_exp_1} (B) shows the size of optimal scale detected by the correction step for each event. The events on bars have small spatial scale size as they represent the correct direction. For the pixels on the square, however, since the bar represents the best flow, the size of window increases as the events get farther away from the bar and a larger scale which would include the bar is needed to find the best direction. This also means what while the optimal scale sizes for the square are symmetric vertically, the presence of the bar means that the horizontal symmetry is lost. The size of the optimal spatial scales are still independent of direction of motion.}

\begin{figure} 
	\begin{center}
		\includegraphics[width=\columnwidth]{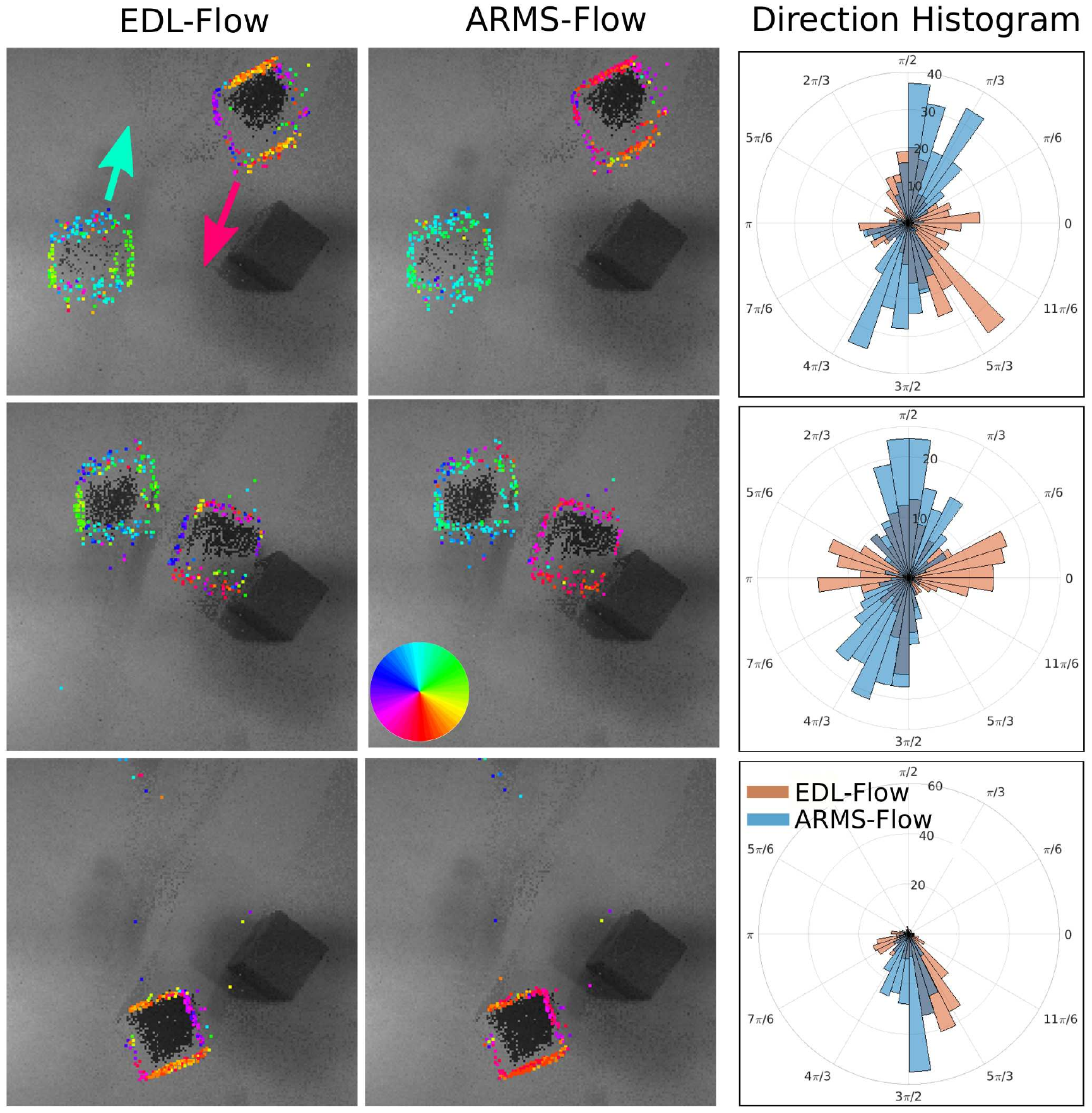}
			\caption{Comparison of EDL and ARMS for two moving objects. The three rows show the direction outputs of the EDL and ARMS flow at three time points. The algorithm works well even when the two objects cross each other closely. Direction histograms show bimodal distribution from ARMS (blue) for the direction of the two individual shapes. The EDL flow (red) however leads to a larger variance and almost a uniform distribution. Even with only one object in the scene (bottom row), EDL flow gives rise to two modalities but ARMS gives a single peak at $3\pi/2$. \label{fig:Fig_results_exp_2}} 
	\end{center}
\end{figure}

%
  
   

\subsubsection{Camera fixed, multiple objects}
Next, we tested the robustness of the multi-scale pooling in case of more than one object moving in front of the camera. To do this, we recorded two simple objects (two squares) moving across the camera in opposite directions. We also have a stationary object in the scene that may lead to noisy events. The experiment shows that the spatial pooling is not affected by multiple objects and the algorithm can find the correct scales for each object independently. Further, when the objects cross each other close by, the algorithm is robust enough to recover the correct directions. Figure \ref{fig:Fig_results_exp_2} shows EDL and ARMS flow output for the two objects and the corresponding direction distribution of events over events in a time window of 100 ms. The left column shows the EDL outputs color coded by the direction of flow estimates. As expected, the estimated directions are normal to the edge directions for each of the objects, leading to an almost uniform distribution of event directions. This, is corrected by ARMS so that we get two distinct peaks in the  direction histograms. As the objects collide and cross each other, the EDL becomes slightly worse and the peaks shift whereas the ARMS flow distribution remains invariant (Figure~\ref{fig:Fig_results_exp_2} [middle row]). Finally, as the objects move further, and we only have one object, the distributions becomes similar to the one in Exp~\ref{Exp1} with single moving object. Again, while EDL gives two peaks for each of the edge orientations, we get a single peak from ARMS indicating the global motion direction.

\begin{figure*}[ht]
  \begin{center}
   \includegraphics[width=1.9\columnwidth]{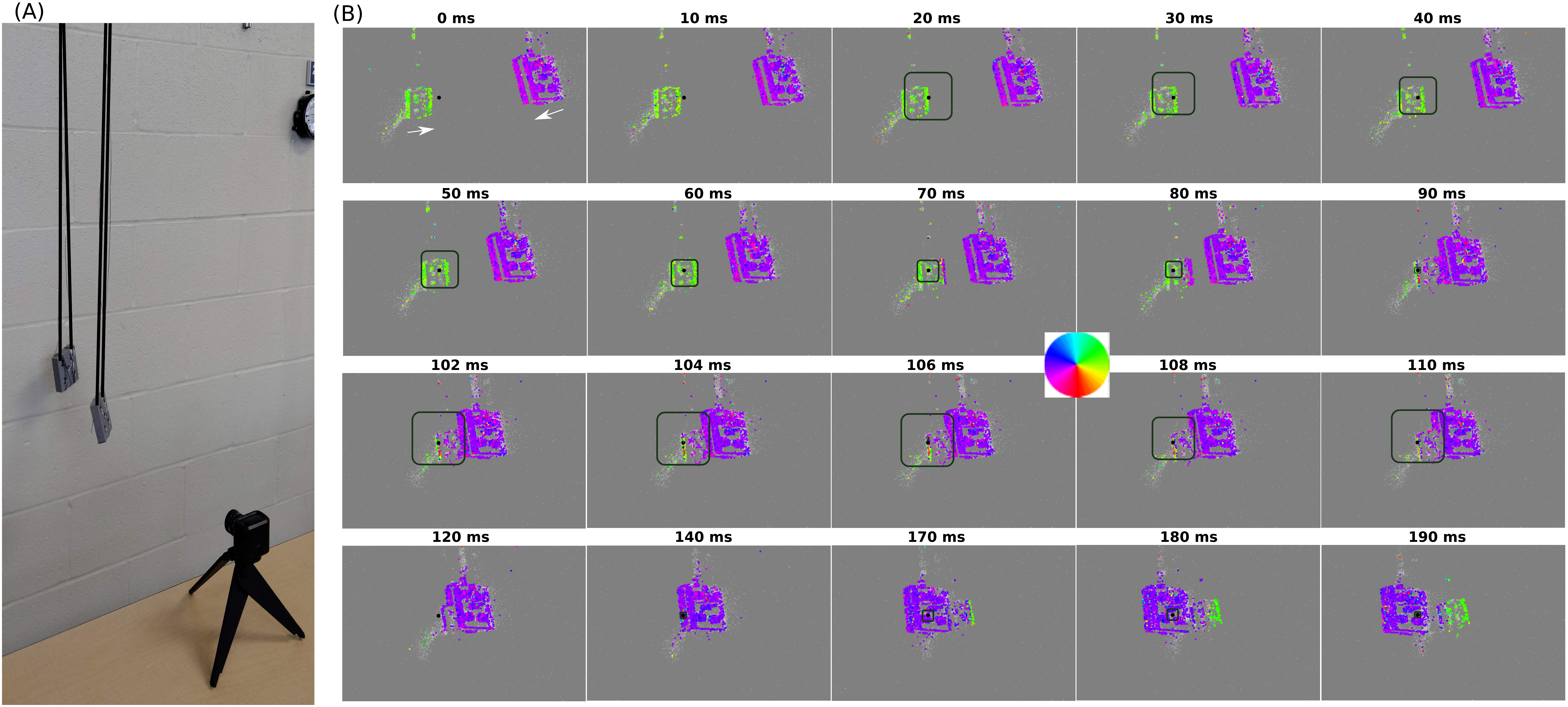} 
	\caption{\revised{Figure shows the setup (A) and ARMS flow directions (B) for a scenario with two pendulums of same length and size but at different depths from the sensor. The figure also shows, as black rectangle, the size of optimal window found by ARMS flow for a given pixel (black dot). We can see that as the two pendulums get close around 70 msec time point, the front edge of the smaller pendulum starts to get erroneous direction estimate due to the higher magnitudes of the closer pendulum. The direction estimates are quickly corrected though, as soon as the two pendulums start to move away (170-190 msec)} \label{fig:double_pendulum}}  
	\end{center}
\end{figure*}

\subsubsection{{Camera fixed, objects occluding}}
{ Next, we tested the robustness of the multi-scale pooling in case when more than one object moving in front of the camera overlap and occlude each other. We setup two pendulums of the same length and size but placed at different depths from the sensor. The pendulums were left to oscillate at out of phase positions such that while both the pendulums are visible to the sensor, there are moments when the two pendulum overlap each other. The setup and qualitative results are shown in figure \ref {fig:double_pendulum}. Figure shows the output of the ARMS flow for a sequence of about 200 msec during which the two objects overlap and then pass each other. \revised{We also show how the optimal spatial scale found by the correction step of the algorithm changes over time as different parts of the objects move over a given pixel represented by the black dot on each panel. The spatial scale is represented by the black rectangle. As the pendulums pass by, the optimal scale selected by the correction step changes depending on which parts of the pendulums are passing through the pixel. The ARMS flow provide the directions close to the actual directions when the pendulums are far apart as in the panels 0 to 10 msec. As the pendulums move closer, the flow direction of the rightmost edge of the smaller pendulum gets corrupted by the higher magnitudes of the larger pendulum. At 102 msec, panel we start to see the impact of the larger pendulum on the spatial scale detection. Even though the obeserved pixel is on the smaller pendulum, the higher magnitudes of the larger pendulum lead to larger spatial scales to be selected as optimal, as represented at 110 msec by the larger rectangle. Finally, as the two pendulums get farther apart at 170 msec, we can see that the flow for the leading edge of the smaller pendulum are quickly corrected and the correct flow values are recovered. } \revised{\sout{We can note that there is an instance when the direction estimate of the farther pendulum is affected by the nearer pendulum's direction but the flow is corrected back within 20 msec as soon as the two pendulums have moved farther away from each other.}} This shows that while the algorithm can get affected instantaneously due to overlapping object, the error remains only for a very short duration of the overlap and can be quickly recovered.}





\subsubsection{Real world scene - Camera mounted on moving car}
\begin{figure*}[ht]
  \begin{center}
   \includegraphics[width=1.5\columnwidth]{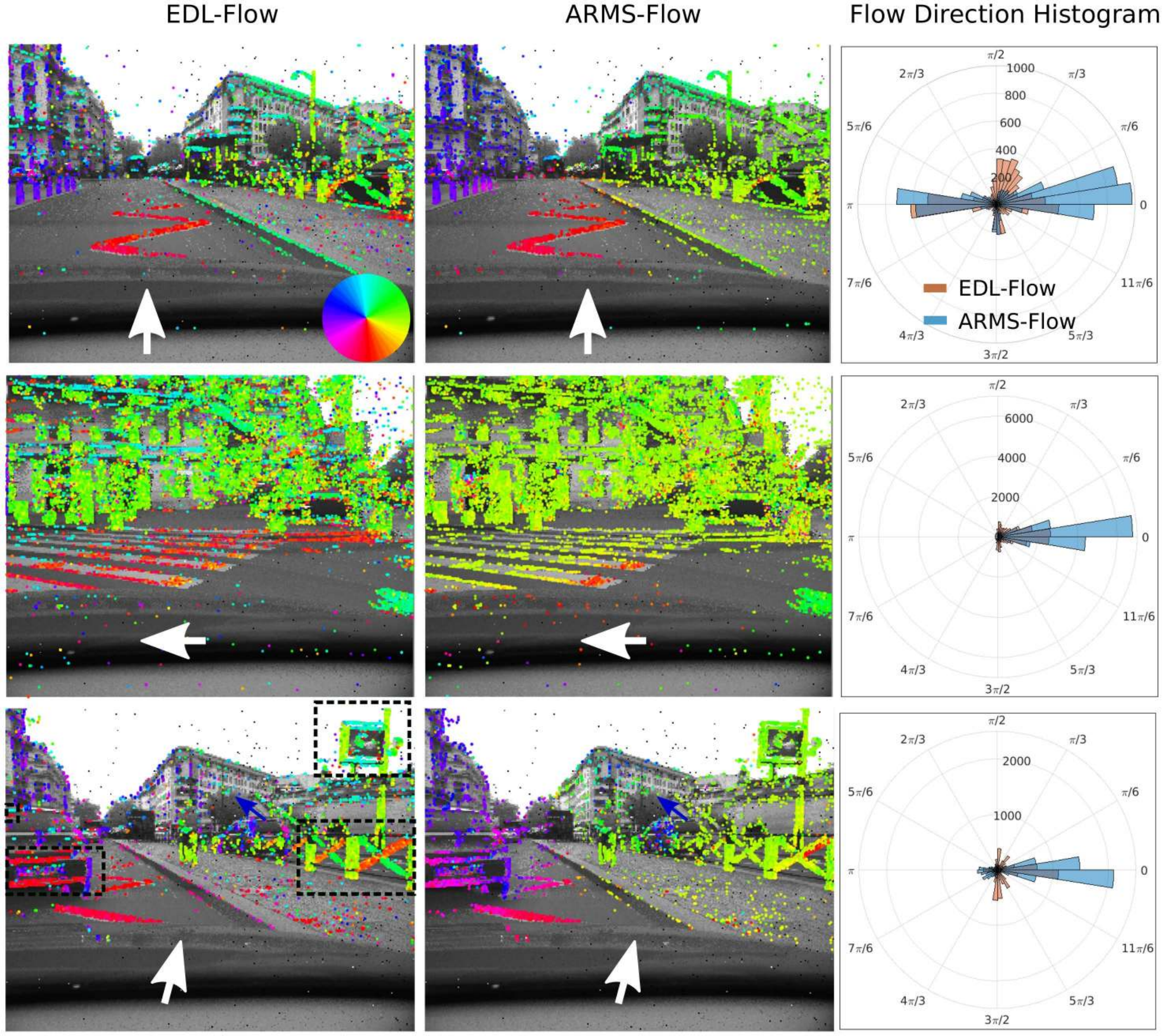} 
	\caption{Figure shows the flow directions for an ATIS mounted on a car moving straight ahead (top), taking a left turn (middle) and navigating around another car (bottom). EDL flow is normal to the edges on most events which is reflected in the histograms by the small incorrect peaks at $\pi/2$ and $3\pi/2$. ARMS-Flow corrects these local abnormalities giving rise to correct direction dependent flow reflected in the two distinct peaks during straight motion and a single large peak around $0~deg$ when the car is turning left. The bottom row provides shows how well the ARMS flow works in a cluttered dynamic case. The black rectangles show the interesting regions in the scene where the normal directions are \revised{improved towards} \sout{\revised{corrected to}} the true global flow while still maintaining the directions of independent moving object like the car on the right which has a relative motion indicated in the forward direction (blue arrow). \label{fig:Fig_results_exp_3}}  
	\end{center}
\end{figure*}

The flow rectification is also assessed through a real world scene in which the event-based camera was mounted on a car moving through traffic along the streets of Paris. The flow obtained from the algorithm corrects the local perpendicular flow to provide a better global flow especially when the car is making turns, where the whole scene should have the same global flow direction. The optical flow corrections can improve the flow directions when the car is turning, making all events predicting the \revised{\sout{correct direction of the global motion}} \revised{apparent global direction of turn}. Further, the combination of speed and flow directions can easily segment objects moving independently from the car. Figure~\ref{fig:Fig_results_exp_3} shows the EDL and ARMS flow for different traffic conditions. The bottom row shows interesting points (marked by black rectangles) in the scene where the ARMS flow successfully corrects erroneous directions of the EDL. These show that the spatial scale estimation works correctly even in a cluttered environment and large motion events. Further, direction estimates of independent moving objects such as cars is not affected by the global motion.

\subsection{DAVIS dataset with ground truth}
\subsubsection{Application to the Event-Camera dataset}
\begin{figure}
  \begin{center}
   \includegraphics[width=1\columnwidth]{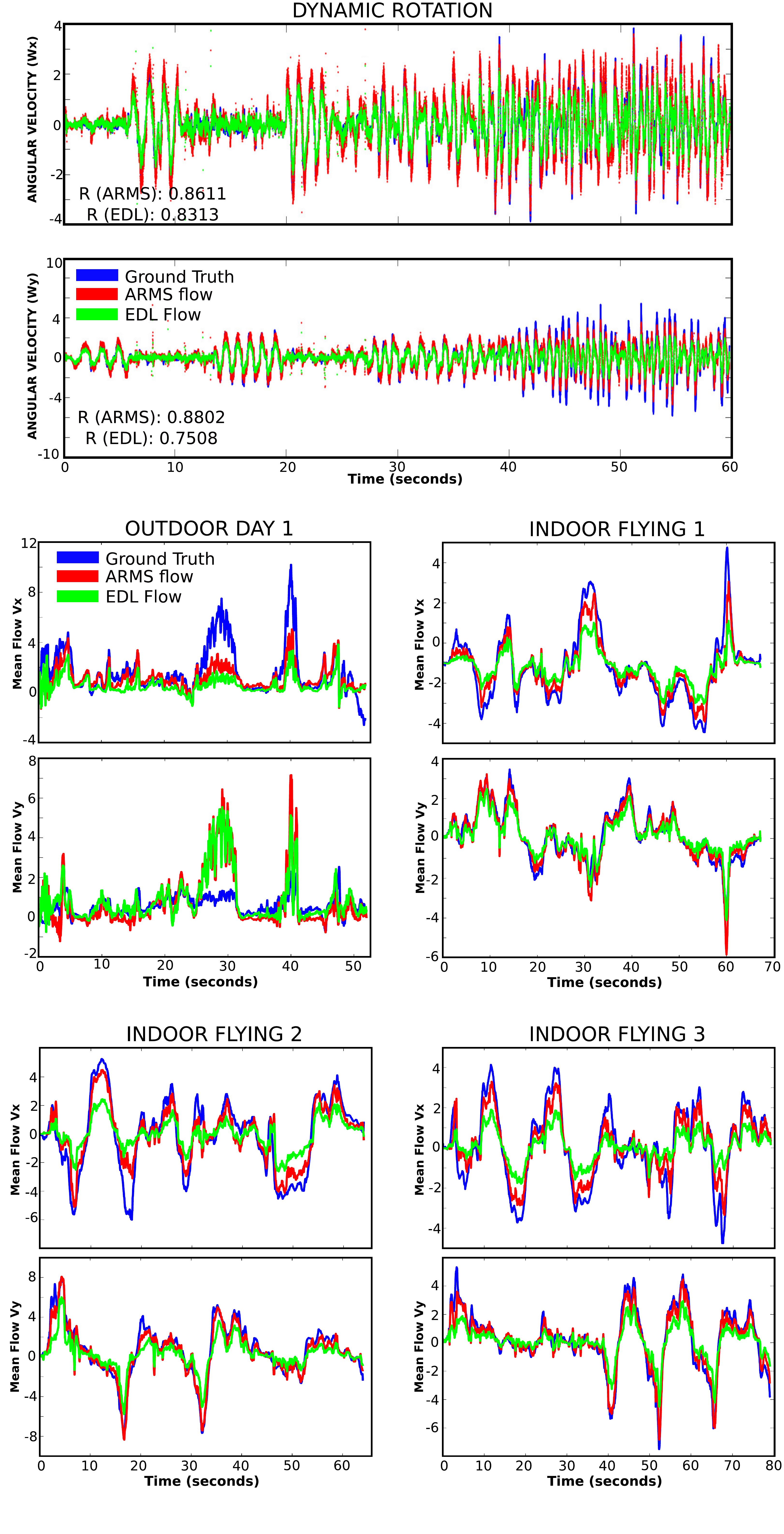}
	\caption{\revised{Comparison of ARMS flow and EDL flow based velocity estimates against the ground truth velocities recorded with an IMU for the rotational dynamics data from DAVIS benchmark data and the Outdoor Driving and Indoor Flying conditions from the MVSEC data over the duration of the recordings.} \label{fig:Fig_results_GT_compare}}  
	\end{center}
\end{figure}

\begin{figure}
  \begin{center}
   \includegraphics[width=1\columnwidth]{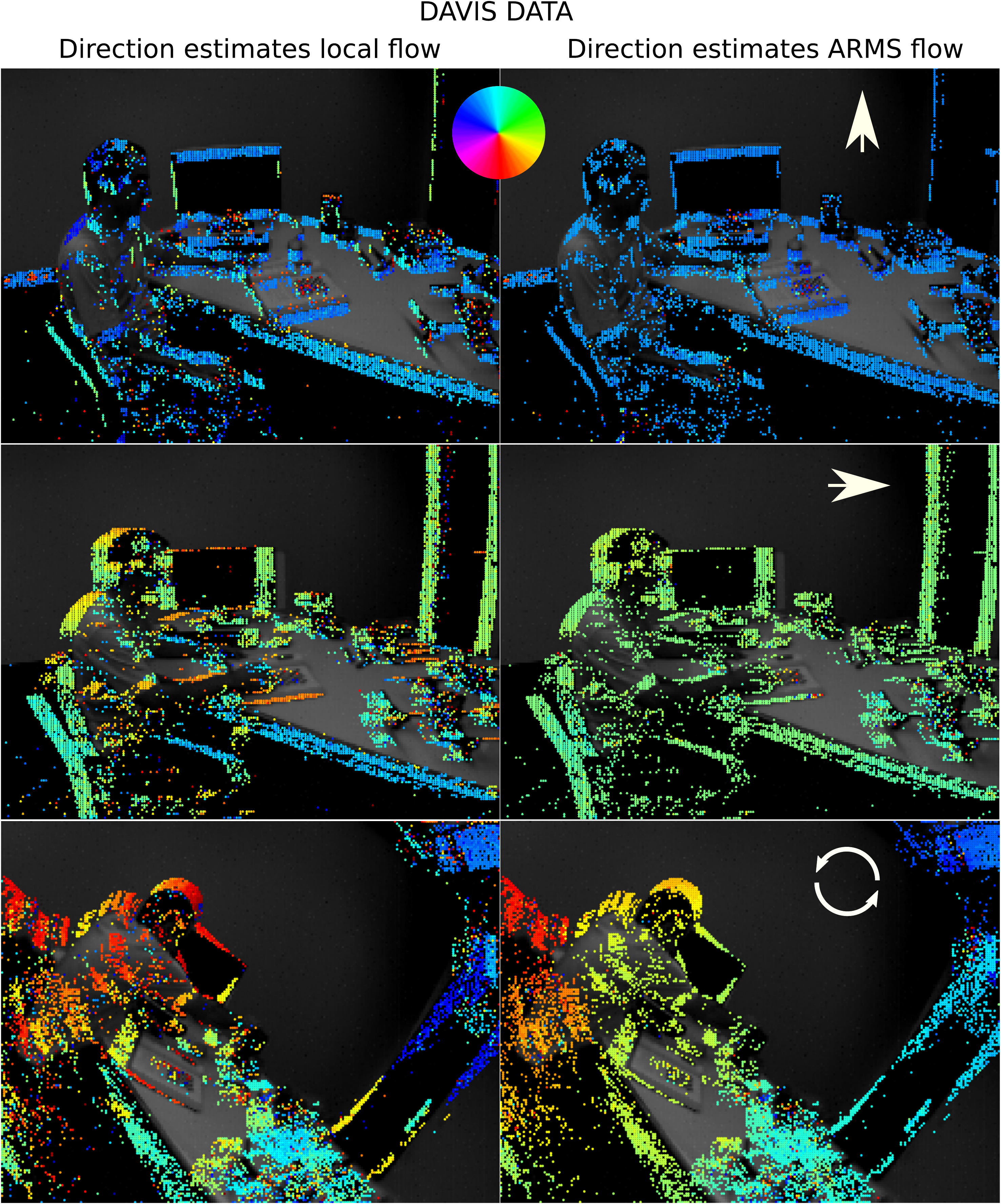}
	\caption{Figure shows the flow results for two different available set of data recorded with DAVIS. \revised{The panels show the EDL and ARMS flow directions computed for data recorded for scenes recorded with camera moving freely while simultaneously recording the events and the motion of the camera with an IMU.}\label{fig:Fig_results_images_DAVIS}}  
	\end{center}
\end{figure}

To test the efficacy of our method on a \revised{\sout{public}} \revised{benchmark} dataset, we chose to implement it on events and images recorded with a DAVIS which also recorded motion of the camera using an inertial measurement unit (IMU) at 1000Hz \cite{Mueggler_2017}. This provides us with not only images of the scene but also the angular velocity of the camera as the scene is recorded. We implemented the ARMS flow on \revised{dynamic rotation scene where an office scene is recorded with camera primarily rotating around its axes. The recording involves different speeds of rotations.} \revised{\sout{two subsets of the dataset - dynamic translation and dynamic 6 degree of freedom The camera recorded and office environment with a person moving around while translating or free moving with increasing speeds.}}.  Figure \ref{fig:Fig_results_images_DAVIS} shows the output flow directions for \revised{the recorded data \sout{these cases}}. Figure \ref{fig:Fig_results_GT_compare} shows the actual imu recordings of the angular velocities and the angular velocities estimated using \revised{the EDL and} our flow. The graphs show \revised{that the predicted velocity can follow the real velocities well as indicated by the high correlation scores which also show improvements over the EDL flow predictions. The flow fails somewhat for the W\_y when the camera moves at higher speeds as seen in the 40 to 60 second mark in bottom graph.} \revised{\sout{high accuracy in these measurements and showing that the method works well in different conditions and cluttered environment with root mean square errors of 0.2003 for dynamic translation and 0.6362 for dynamic 6-dof motion. }}

\begin{figure}
  \begin{center}
  \includegraphics[width=0.95\columnwidth]{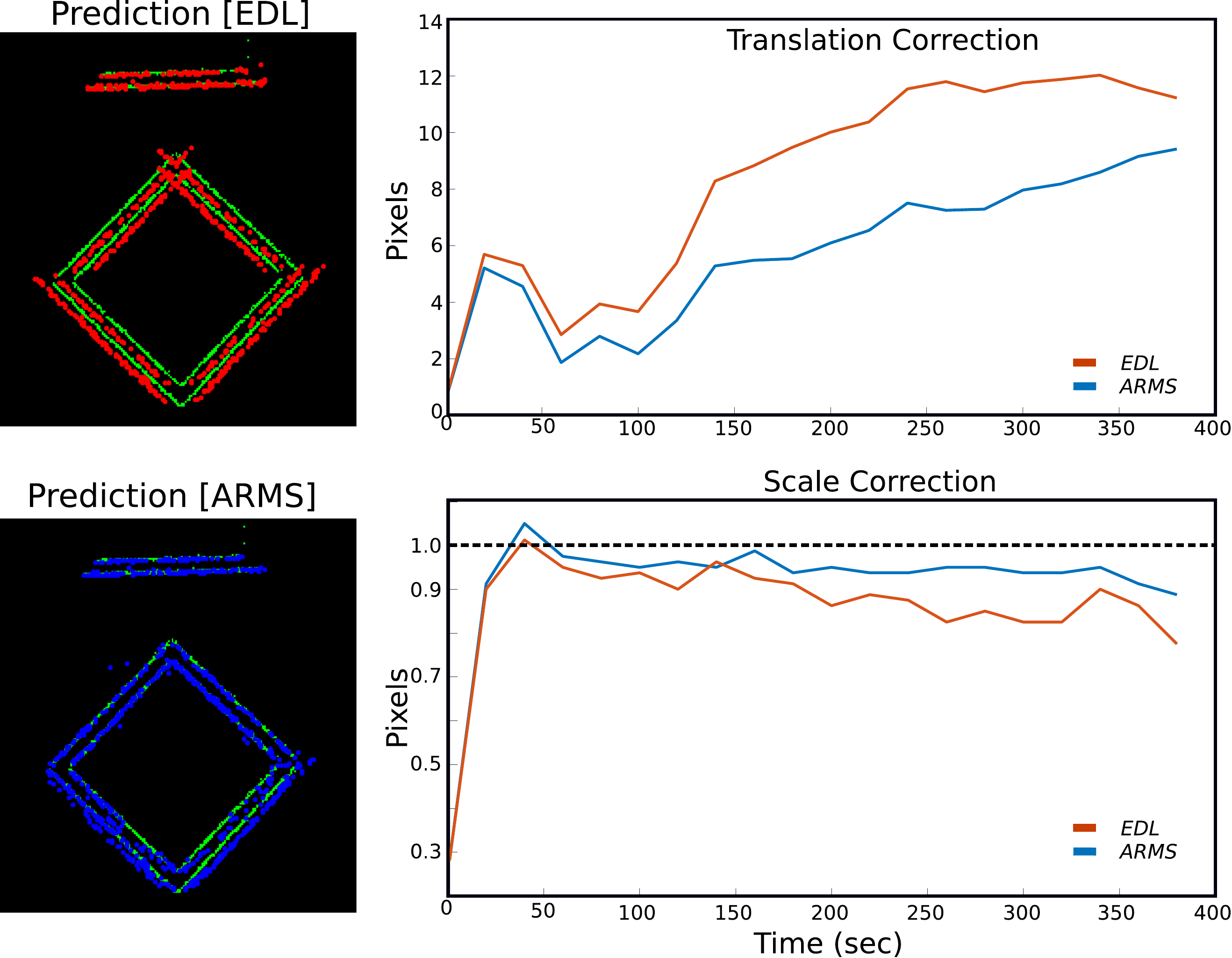}  
\caption{The figure shows the predicted events based on EDL and ARMS flow at 250ms in future. Green dots indicate the actual future events while the red and blue dots indicate events predicted by the two flows. The scaling and translation error show how well the ARMS flow keeps the affinity of the object events. The ARMS flow has required scaling closer to 1 and translation error lower than the EDL flow error indicating that all events point to the true direction of motion.  \label{fig:Fig_results_exp_1_pred}}
  \end{center}
\end{figure}

\begin{figure*}
  \begin{center}
   \includegraphics[width=1\linewidth]{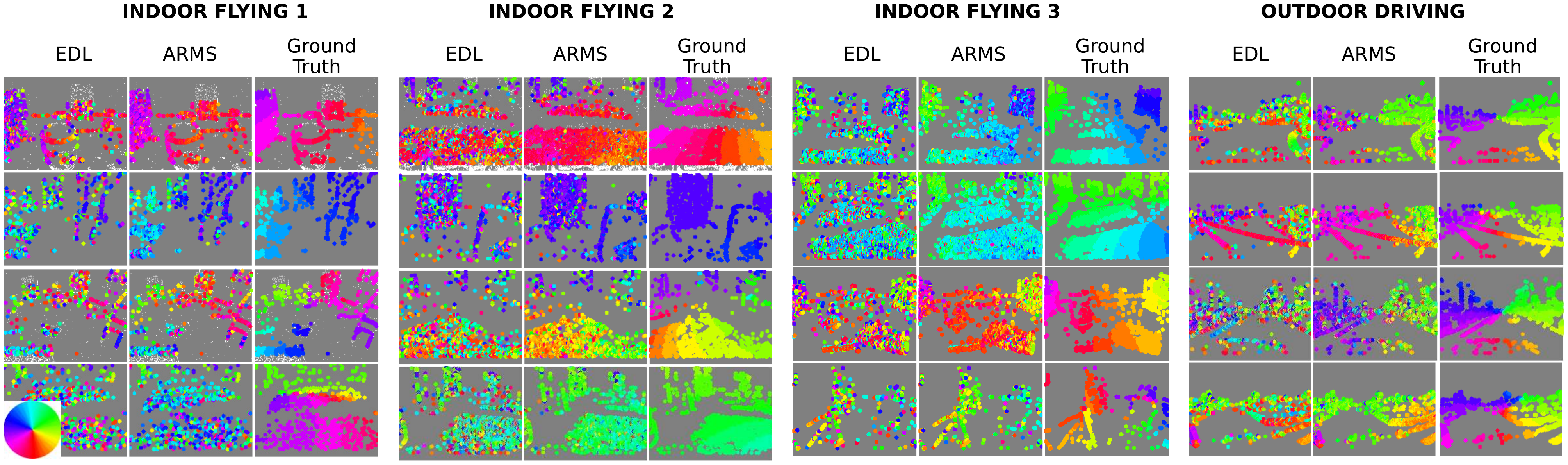}
	\caption{\revised{Figure shows the flow results for events recorded from different conditions from the MVSEC benchmark data. The panels show flow directions from EDL and ARMS flow along with the ground truth directions. The ARMS flow vastly improves the EDL flow estimates and generally is close to the ground truth directions. We also show examples when the ARMS flow fails as shown in the bottom panels for Indoor flying 1, Indoor flying 3 and Outdoor Driving conditions.} \label{fig:Fig_results_images_MVSEC}}  
	\end{center}
\end{figure*}

\subsubsection{Application to MVSEC dataset}
Finally, we also have applied the optical flow rectification algorithm on the MVSEC dataset~\cite{EVFlowNet} as shown in Figure \ref{fig:Fig_results_images_MVSEC}\revised{\sout{ (bottom panel)}}. \revised{Fig. \ref{fig:Fig_results_GT_compare} shows the ground truth velocities over the duration of recordings for the several conditions and the corresponding mean EDL amd ARMS flow. The MVSEC dataset ground truth combines several sensor data along with the framed images obtained from the camera to create dense flow for every image taken by the camera. Since our algorithm produces flow for every event, to compare with the ground truth, for every pixel we averaged the flow produced between the two images. Figure \ref{fig:Fig_results_images_MVSEC} shows qualitative comparisons between the EDL, ARMS and ground truth directions from snapshots of the data taken in different conditions. We find that the ARMS flow vastly improves the EDL flow and generally performs well in different conditions. There are also conditions where the flow does perform poorly as shown in the bottom panels for the conditions - indoor flying 1, indoor flying 3 and outdoor driving. Such situations arise when the camera/scene are moving too slow for the EDL flow to provide good flow outputs or when the direction changes are too large for the flow to correct quickly creating discontinuity in the event plane. We must also note that the ground truth is built from the information of a LIDAR, (fused with the GPS and he IMU), which provides depth information within a range of 100m in the case of outdoor driving sequences. On the other hand, the ARMS flow is estimated only from a single camera with much lower spatial resolution. This leads to errors in the flow when the car is both turning and moving forward as shown in the bottom panel of the outdoor driving sequence in Fig. \ref{fig:Fig_results_images_MVSEC}. The ARMS flow seems to "see" only the dominant apparent motion of a left-to-right translation whereas the ground truth shows an expanding flow due to far structures in the scene.} To further quantify the performance, \revised{as used in \cite{EVFlowNet},} we computed the average endpoint error (AEE) $ = \sum{||(\hat{V} - V_{true})||_2}$ where $V_{true}$ is the ground truth derived from the dataset and $\hat{V}$ is the computed flow. The AEE was only computed over events rather than whole images. The performance was compared against the state of the art algorithms – EV-flownet \cite{EVFlowNet} and Event based visual flow (EV-flow) \cite{Benosman2014}. The results of the quantification are presented in Table \ref{table: results}. The errors for the Indoor Flying conditions ($In\_Fly$) were taken from \cite{EVFlowNet}. We additionally report our error estimates for the outdoor driving conditions \revised{for which the error values were not provided with the dataset.}  


\begin{table}[ht]
\begin{center}
\begin{tabular*}{\columnwidth}{@{\extracolsep{\fill}}|c|c|c|c|c|c|}
\hline
Method & In\_Fly1 & In\_Fly2 & In\_Fly3 & {Out\_Day} & {Out\_Night} \\ 
\hline
EV-flow & 1.03 & 1.72 & 1.53 & {N/A}  & {N/A} \\
Net &  &  &  & &  \\
\hline
EV-flow & 2.45 & 2.42 & 5.35 & {3.87} & {5.53} \\
{(EDL)} &  &  &  & &  \\
\hline
ARMS & 1.52 & 1.59 & 1.89 & {2.75} & {4.47} \\
flow &  &  &  & &  \\
\hline
\end{tabular*}
\end{center}
\caption{Average Endpoint Error (AEE) in pixel, for \& {five}  MVSEC data set. The EV FLow-Net paper does not provide any error performance for the outdoor sequences. \label{table: results}}
\end{table}

The proposed method shows remarkable improvements over the EV-flow and even though the algorithm is simple and works only on events, its performance matches that of the EV-flownet which requires elaborate learning network and is trained using both events and grayscale images. This means that to use it, one must have both event and image recordings for training of new scenes. Our method on the other hand only uses change events. We think that retraining the EV-flownet on binary images or only on events, or adding additional grayscale information into our algorithm would be a more suitable comparison and should close the performance gap between the two algorithms.

\begin{figure}
  \begin{center}
  
   \includegraphics[width=1\columnwidth]{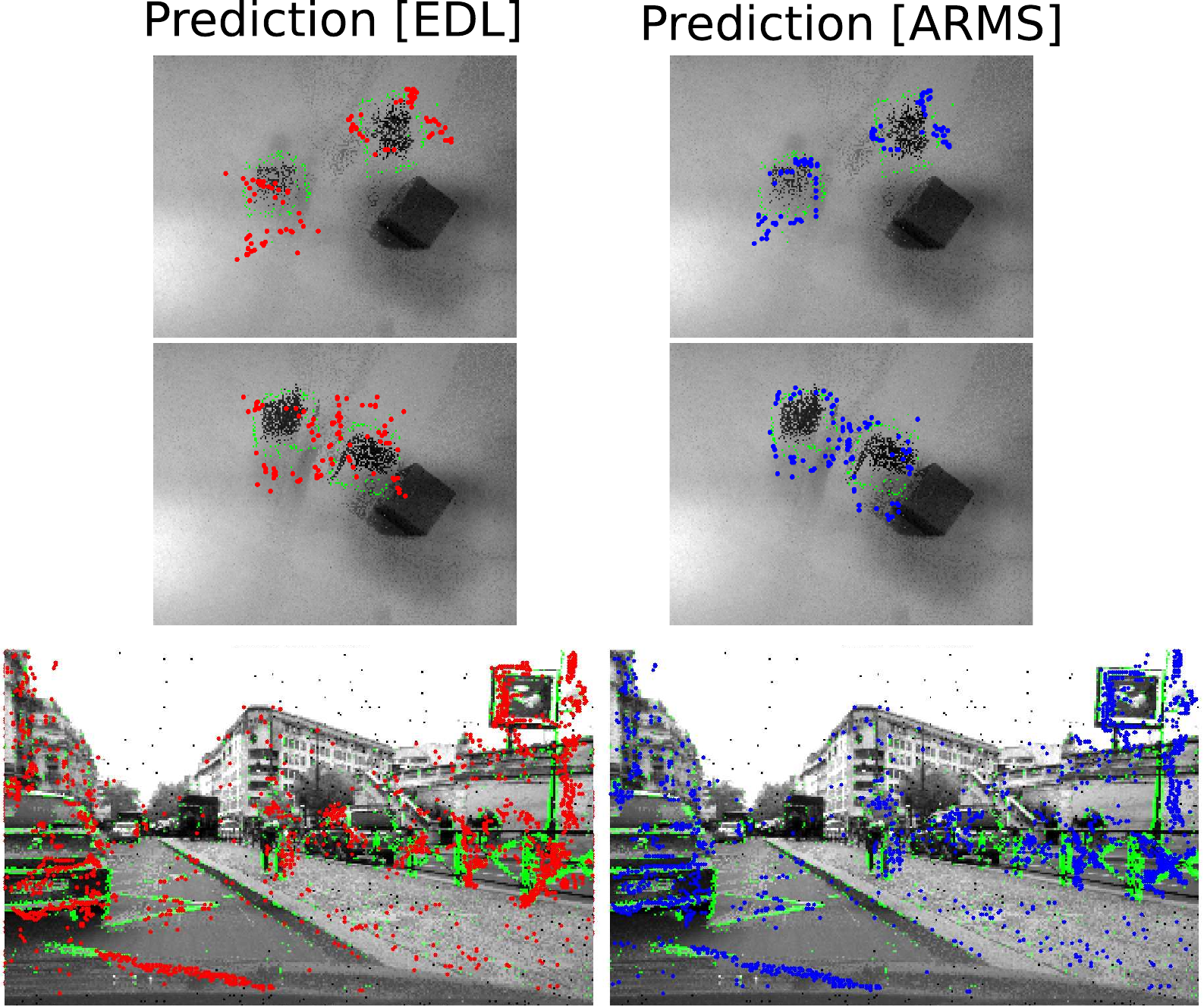}  
    
	\caption{Figure shows the performance of EDL and ARMS flow on prediction of events for the shapes and moving car scenario. The images show the actual events (green), the predicted events using EDL (red) and the predicted events using ARMS (blue). The figures show that the ARMS flow can greatly improve the prediction in both clean and cluttered and complicated environment invariant to the shapes or number of objects. \label{fig:Fig_results_exp_8}
}  
  \end{center}
\end{figure}

\subsection{Event based prediction using ARMS flow}

\subsubsection{Trivial Case}

The corrected direction estimates using ARMS flow can greatly improve the prediction of rigid object over traditional plane fitting methods. 
Figure \ref{fig:Fig_results_exp_1_pred} shows the actual future events (green) and the predicted events for EDL (red) and ARMS (blue) flow using events that occurred 250 msec in the past. The figure shows that using ARMS flow, all the predicted events of the square form another square but if the directions are not the same as in case of the EDL, the predicted shape is not rigid anymore and does not form a square. To quantify the performance of the two flows, we compute how well the predicted events from local and corrected flow maintain the rigidness of the object. That is, we compute the affine transformation needed to map the predicted events to the actual events. To simplify, we assume zero rotation and perform only translation and scaling. The graphs in Figure~\ref{fig:Fig_results_exp_1_pred} show the scaling and translation needed for the EDL and ARMS flow for a sequence of 360 msec broken into event clusters of 20 msec each. A perfect prediction would imply no scaling (i.e. scaling correction = 1) and no translation (translation correction = 0). The mean translation error for ARMS flow was $6.52$ pixels per event vs $8.70$ pixels per event for EDL flow. More importantly the scaling error in ARMS was only $0.085$ compared to $0.141$ in case of EDL. These results show that our proposed ARMS flow reduces the translation error and requires almost no scaling corrections showing that this flow can be used successfully to perform predictions on moving rigid objects.\\

\subsubsection{Prediction of multiple objects and moving car}
Next, we perform event by event predictions for the multiple moving objects and moving car scenarios mentioned in the previous section. Figure~\ref{fig:Fig_results_exp_8} provides the prediction results from the non rectified and the rectified flow for these sequences. Contrary to the two previous sequences, as the scale from the scene is not easily extracted, we do not assess the impact of the prediction by measuring the affine deformation parameters and show only the prediction results. We managed thanks to the rectified flow, to predict events up to 250 msec. The ARMS flow can predict the event-by-event locations even in such a highly dynamic scene across all directions and again helps to maintain the affinity of different objects, such as the car, the person walking and the environment such as the dividers, poles and signs.\\

\begin{figure}
    \centering
    \includegraphics[width=0.9\columnwidth]{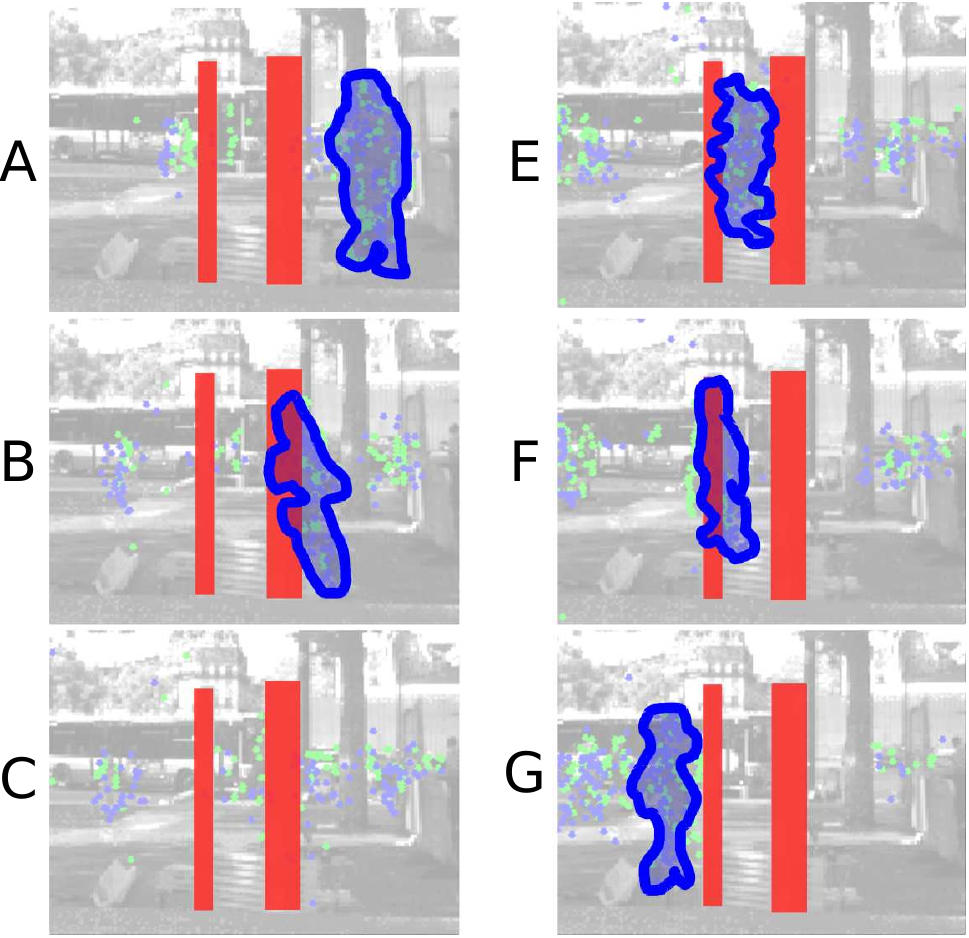}
    \caption{Predictions of the motion of a person up to 200ms using ARMS flow in real world cluttered environment occluded by objects. The images show predictions when the pedestrian passes behind two poles (highlighted by red masks). The location of the pedestrian is masked (blue) by clustering the predicted event location and creating a single blob.
    }
    \label{fig:occlusion}
\end{figure}

\begin{figure*}[ht]
  \begin{center}
  
   \includegraphics[width=1.6\columnwidth]{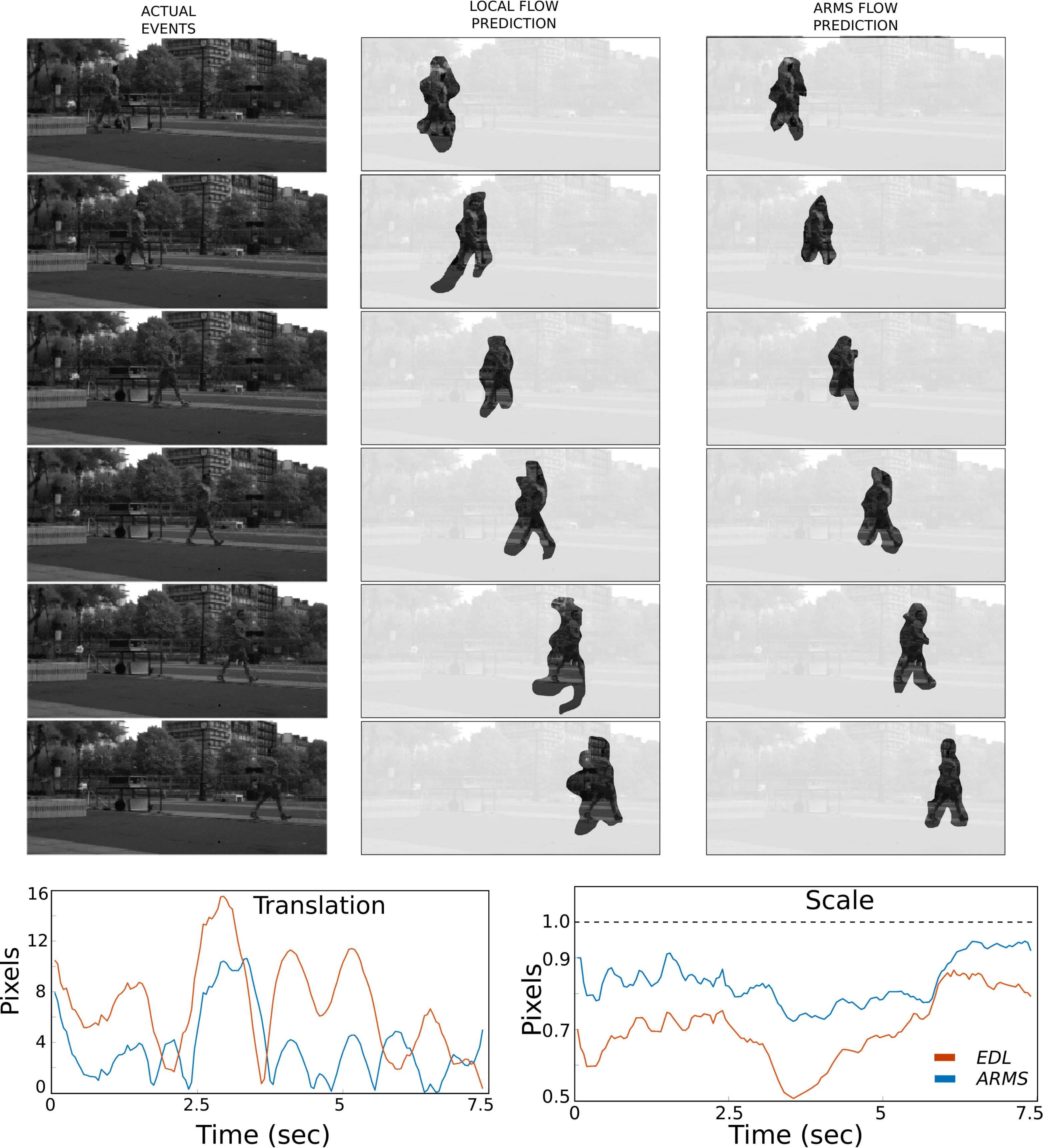}  
    
	\caption{Figure shows the possibility of performing predictions based on optical flow estimates for people passing by on a street at two different points in time. The images shows the actual events (green), predicted events using EDL (red) and predicted events using ARMS (blue) over grayscale images obtained from ATIS. The predictions from ARMS clearly show much better fit with the actual events while the predictions from EDL tend to create inflated shapes. This is quantified in the graphs showing the transformation required to fit the predicted events to the true events. The EDL required larger corrections in both scaling and translation compared to ARMS predictions. \label{fig:Fig_results_exp_4}
}  
  \end{center}
\end{figure*}

\subsubsection{Prediction in cluttered scene with occlusions}
We also computed the flow and performed predictions at 200ms in a more cluttered scenario where a pedestrian was tracked while passing behind stationary objects and across other pedestrians in a busy street scene. The results are presented in Figure \ref{fig:occlusion}. The predictions very clearly match the actual events. This example is a typical showcase for the robustness of the prediction to occluding perturbations. We do not need to resort to more complex procedure such as some Kalman filtering to achieve accurate prediction.

\subsubsection{Prediction of moving pedestrian on street}
To further measure the prediction capabilities using our flow method we placed the ATIS on a street corner and recorded pedestrians passing by. As in the trivial case example, for each incoming event, we make prediction on where the event will occur after 500 msec using the optical flow computed with both the local plane fitting and the new ARMS flow algorithm. We performed the transformation estimation for event clusters over 50 msec  time windows. While the direction component of the flow for each event is used as it is, the speed component of the flow is normalized by the mean speed i.e., each event $i$ in the event-cluster has speed $M$ and direction $\theta_i$, where $M$ is the mean speed of all events and $\theta_i$ is individual flow directions. Using these predictions, a reconstruction of the motion is made based on local and corrected flow as shown in Figure \ref{fig:Fig_results_exp_4} by red and blue dots respectively. The figure shows that the ARMS flow can predict the position of the man up to next 500 msec very accurately. We used the transformation metric as used in the previous experiment to compare the performance of the two methods. The graphs show that the ARMS method outperforms the EDL through the sequence of recording for both scaling and translation corrections. The mean translation error for EDL was $7.1240$ pixels while that for ARMS was $4.7558$ pixels while the scaling error was $0.297$ and $0.167$ for EDL and ARMS respectively. Qualitatively, the cluster formed by the predictions based on the ARMS-Flow is less noisy and more compact and is much closer to the real events. This shows that our algorithm can maintain the shape on a rigid moving object even when the predictions are made on an event by event basis and therefore at very high temporal rates.\\\\



\begin{table}[h!]
\centering
  \begin{tabular*}{1\columnwidth}{@{\extracolsep{\fill}}|c|c|c|c|c|}
  \hline 
{\textbf{Data} } & {\textbf{Num}} & {\textbf{Actual}} & {\textbf{Compute}} & {\textbf{Rate}} \\
{\textbf{} } & {\textbf{Events}} & {\textbf{duration(sec)}} & {\textbf{time(sec)}} & {\textbf{(Evn/sec)}}\\
      \hline 
     
      \hline 
      Shapes & 111999 & 0.678 & 0.845 & 132510 \\
      \hline 
      BarSquare & 1.25e6 & 5.8 & 8 & 156305 \\
      \hline
      OutDay 1 & 5e6 & 13.41 & 25.9 & 132926 \\
      \hline 
      OutNight 1 & 5e6 & 17.2  & 23.3 & 114286 \\
      \hline 
      IndoorFly 1 & 5e6 & 29.87 & 28.2 & 192205 \\
      \hline
      IndoorFly 2 & 5e6 & 19.58  & 28.16 & 127500 \\
      \hline 
      IndoorFly 3 & 5e6 & 24.21 & 26 & 172308 \\
      \hline

  \end{tabular*}
  \caption{\revised{Benchmark computation times on an Intel E5-1603 processor.} \label{table: benchmark}}
\end{table}

     

\section{Conclusion}
Event driven sensors provide an efficient sampling method to solve computer vision problems with scope for developing novel algorithms in temporal domain. Optical flow is an important feature for most vision based open problems and estimating fast yet robust flow is a crucial step. While some interesting algorithms have been developed to estimate visual flow using the event-driven sensors, they either fail to solve the aperture problem due to the emphasis on local spatio-temporal computation or are inefficient and do not really use the high event speeds of these sensors. \\
In this paper, we have presented a novel visual flow algorithm that not only solves the aperture problem but also performs on an event-by-event basis justifying the use of event-driven sensors. In fact, we exploit the intrinsic property of the event based optical flow algorithm, that allows for correcting the directions of erroneous local flow estimates. We have shown here that the algorithm works in real world scenarios, in case of both stationary and moving camera. The algorithm is invariant to the number of objects or their size and does not require additional processing steps such as object detection and tracking. This fast implementation allows us to perform truly event based prediction of moving objects from 250 to 500 msec in future without affecting the shape and size of the object. This is equivalent to making estimations of position of an object upto 10-25 frames in future when using a traditional frame-based camera. To the best of our knowledge, we could not find any methods using event-driven sensors that have attempted to perform such accurate predictions without any temporal binning of events. Further, these predictions are invariant to the size and number of independent objects in the scene. These predictions can allow higher order recognition and tracking layers to perform at the high temporal rates at which events are generated. Our future goals are to use this algorithm as part of an autonomous driving car sensor system to allow for fast collision detection and detect abnormal driver and pedestrian behavior. {Further, the spatial scaling method works not only for plane fitting based local flow methods but any local flow methods that satisfy the condition that the flow magnitude is related to the contour of the edge or object in motion.}
\\
\revised{\sout{Our algorithm does have certain short comings, as it only corrects attempts to correct the direction of individual flow but does not perfectly correct the magnitude of the flow.}} \revised{ Since, our algorithm does not use higher order features, the true flow is only possible if an edge with true direction motion is present. In general, the flow improves depending on the presence of edges that are close to being orthogonal to the true flow direction. Also, since the flow is provided by the mean of local flows in a spatial window, the estimated flow is slightly moved away from the true flow.} An improved prediction of the magnitude of flow could allow us to make predictions at even longer duration of up to a few seconds. A possible solution to improve the estimates could be by using the gray scale information provided by an ATIS like neuromorphic sensor. \revised{Analysing MVSEC and DAVIS data also show that ARMS flow can fail in certain scenarios especially in outdoor conditions where objects are far and the events on the camera plane themselves are not enough to compensate for the depth of the objects. This means that in cases when the car is making turns while moving forward the ARMS flow can only see the apparent motion of the whole scene moving whereas the depth information using a LIDAR would allow measure the expanding flow. In indoor flying cases, we found that while ARMS flow can remarkably improve the EDL flow output, errors occur if the drone makes large, sudden change in directions as this leads to discontinuity in the event plane and leads to large errors in local flow computation itself. The ARMS flow still performs close to ground truth for most of the duration of the flight sequences.} {Many new techniques for motion correction use contrast maximization methods \cite{Stoffregen_2019_ICCV} to segment events from object moving at different directions and speeds. This may allow for flow to be computed on events of individual objects. This could also improve the magnitude estimations but the events correction using this technique is still to some extent affected by the aperture problem \cite{Stoffregen_2019}.} 
\\
In terms of the memory and CPU requirements, the algorithm was implemented in C++ running on single core Intel E5-1603 processor, achieving on average a \revised{computation} \revised{\sout{processing}} rate of \revised{120 KEvents/second \sout{2.5 kevents/seconds} [Table \ref{table: benchmark}]} and requires very small amount of memory that increases linearly with the pixels resolution of the sensor. While the traditional CPU is enough for real-time processing on a qVGA sensor, a parallel neuromorphic hardware implementation could make the algorithm independent of the sensor resolution and allow real time motion based visual processing for larger sensor arrays.   



\clearpage



%

%
\begin{IEEEbiography}
[{\includegraphics[width=1in,height=1.25in,clip,keepaspectratio]{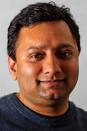}}]{Himanshu Akolkar is  currently  a Post Doctoral Associate  at the University of Pittsburgh. He received his M.Tech. degree from IIT, Kanpur (India) in EE  and PhD from IIT, Genoa (Italy) in Robotics after which he had a Post Doctoral stint at Université Piérre ét Marie Curie. His primary interest is to understand the neural basis of sensory and motor control to develop an intelligent machine.}
\end{IEEEbiography}

\enlargethispage{5in} 
\begin{IEEEbiography}
[{\includegraphics[width=1in,height=1.25in,clip,keepaspectratio]{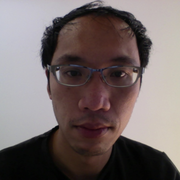}}]{Sio-Hoi Ieng is currently  an  Associate  Professor with   Sorbonne Université,   Paris,   France,   and   a member   of   the   Vision   Institute,   Paris.   He   was involved  in  the  geometric  modeling  of  catadioptric and   non-central   vision   sensors   and   their   link   to the  caustic  surface.  His  current  research  interests include  neuromorphic  and  event-based  vision  perception algorithms and computer vision, with a special reference to the understanding  of general visual sensors,  exploring  cameras  networks,  and  studying the connection between biologic and artificial vision.}
\end{IEEEbiography}

\begin{IEEEbiography}
[{\includegraphics[width=1in,height=1.25in,clip,keepaspectratio]{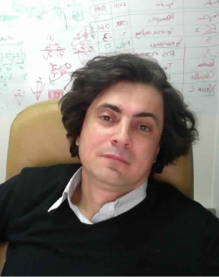}}]{Ryad   Benosman received  the  M.Sc.  and  Ph.D. degrees  in  applied  mathematics  and  robotics  from University Pierre and Marie Curie in 1994 and 1999, respectively. He is Associate Professor with University Pierre and Marie Curie, Paris, France, leading the Natural Computation and Neuromorphic Vision Laboratory, Vision Institute, Paris. His work covers neuromorphic  visual  computation  and  sensing.  He is currently involved in the French retina prosthetics project  and  in  the  development  of  retina  implants and cofounder of Pixium Vision a french prosthetics company. He is an expert in complex perception systems, which embraces the conception, design, and use of different vision sensors covering omni-directional 360 degree wide-field of view cameras, variant scale sensors, and non-central sensors. He is among the pioneers of the domain of omni- directional vision and unusual cameras and still active in this domain. He has been involved in several national and European robotics projects, mainly in the design of artifcial visual loops and sensors. His current research interests include the understanding  of  the  computation  operated  along  the  visual  systems  areas and establishing a link between computational and biological vision. Ryad Benosman has authored more than 100 scientific publications and holds several patents  in  the  area  of  vision,  robotics  and  image  sensing.  In  2013, he  was awarded with the national best French scientific paper by the publication LaRecherche for his work on neuromorphic retinas}
\end{IEEEbiography}



\vfill


\end{document}